\documentclass[runningheads]{llncs}
\usepackage[T1]{fontenc}
\usepackage{graphicx,verbatim}

\usepackage{etoolbox}
\usepackage{booktabs}
\usepackage{adjustbox}
\usepackage{multirow}
\usepackage{lipsum}
\usepackage{color}
\usepackage{amsmath}

\newcommand\blfootnote[1]{%
  \begingroup
  \renewcommand\thefootnote{}\footnote{#1}%
  \addtocounter{footnote}{-1}%
  \endgroup
}

\begin{document}
\title{Pathology~Report~Generation~and Multimodal~Representation~Learning for~Cutaneous~Melanocytic~Lesions}
\author{
Ruben T. Lucassen\inst{1,2, *} \and
Sander P.J. Moonemans\inst{3, *} \and
Tijn van de Luijtgaarden\inst{3} \and
Gerben E. Breimer\inst{1} \and
Willeke A.M. Blokx\inst{1} \and
Mitko Veta\inst{2,4}
\authorrunning{R.T. Lucassen et al.}
\institute{Dept. of Pathology, University Medical Center Utrecht, Utrecht, the Netherlands\\\email{r.t.lucassen@umcutrecht.nl}\and Dept. of Biomedical Engineering, Eindhoven University of Technology,\\Eindhoven, the Netherlands\and Dept. of Mathematics and Computer Science, Eindhoven University of Technology,\\Eindhoven, the Netherlands\and Eindhoven Artificial Intelligence Systems Institute (EAISI),\\Eindhoven University of Technology, Eindhoven, the Netherlands}}

\maketitle 
\begin{abstract}
Millions of melanocytic skin lesions are examined by pathologists each year, the majority of which concern common nevi (i.e., ordinary moles). While most of these lesions can be diagnosed in seconds, writing the corresponding pathology report is much more time-consuming. Automating part of the report writing could, therefore, alleviate the increasing workload of pathologists. In this work, we develop a vision-language model specifically for the pathology domain of cutaneous melanocytic lesions. The model follows the Contrastive Captioner framework and was trained and evaluated using a melanocytic lesion dataset of 42,512 H\&E-stained whole slide images and 19,645 corresponding pathology reports. Our results show that the quality scores of model-generated reports were on par with pathologist-written reports for common nevi, assessed by an expert pathologist in a reader study. While report generation revealed to be more difficult for rare melanocytic lesion subtypes, the cross-modal retrieval performance for these cases was considerably better.

\setcounter{footnote}{0}
\blfootnote{* R.T. Lucassen and S.P.J. Moonemans are co-first authors.}
\end{abstract}

\section{Introduction}
Millions of moles\,\footnote{Based on the ratio of mole to melanoma excisions, which is around 20\,:\,1~\cite{lott2018population,lucassen2024artificial}, and the estimated global melanoma incidence of 332 thousand cases in 2022~\cite{bray2024global}.} are surgically removed by general practitioners and dermatologists each year. After removal, these melanocytic lesions are examined by pathologists to exclude cutaneous melanoma.
Despite it being the most deadly form of skin cancer, melanoma has a relatively low prevalence, accounting for roughly 5\% of the excised melanocytic lesions~\cite{lott2018population,lucassen2024artificial}.
In contrast, the large majority of melanocytic lesions, known as common nevi, are often clearly benign based on the histology and straightforward to diagnose.
While it may only take an experienced pathologist a few seconds to diagnose most common nevi, each case requires a written pathology report that describes the visual features in detail, which can easily take up to several minutes to write and is a fairly repetitive task. 
At the same time, pathologists' workload is already high and estimated to increase even further~\cite{vanderLaak2021,metter2019trends}. Hence, automating part of the pathology report writing can potentially reduce the work pressure on pathologists~\cite{berbis2023computational}.

Recent advancements in deep learning have focused on the integration of multiple information modalities, such as images and text~\cite{li2023blip,liu2024visual,radford2021learning,yu2022coca}. Some of the first vision-language models applied in the pathology domain, including PLIP~\cite{huang2023visual} and CONCH~\cite{lu2024visual}, used histopathology images with corresponding captions from social media or scientific literature. Unable to scale to the large amount of visual data in whole slide images (WSIs), these first models were superseded by vision-language models such as PRISM~\cite{shaikovski2024prism}, PathAlign~\cite{ahmed2024pathalign}, and TITAN~\cite{ding2024multimodal}. Although these models have impressively demonstrated the capability to generate text consistent with visual features from entire hematoxylin and eosin (H\&E)-stained WSIs, so far, this text has mainly been limited to a single sentence describing the diagnosis and listing some key findings. A complete pathology report, however, typically also contains a more extensive description of all visual characteristics. It is this part of the report that is most time-consuming and repetitive to write and, therefore, what pathologists would likely benefit from most if automated. 

A central part of the current vision-language models is the alignment of visual and textual information in the same feature space, enabling both uni- and cross-modal comparison of feature vectors. As a secondary use case, these models could be leveraged for content-based retrieval of cases from databases at pathology departments with a digital workflow. For example, this could enable pathologists to retrieve cases with a specific visual feature, which may not be described in the corresponding reports, based on a text prompt.

The contributions of this work are threefold: (1) We develop a vision-language model for the specific pathology domain of cutaneous melanocytic lesions, which for many pathology departments constitutes a substantial part of the complete case load. Similar to how pathology reports are written at case level, the models were trained using image data from all H\&E-stained WSIs of a single case at once. Moreover, the pathology reports were preprocessed to include both diagnoses and detailed descriptions of the visual features, while excluding information that cannot be inferred from the H\&E-stained WSIs to prevent hallucination\,\cite{ji2023survey}. (2) We evaluate the accuracy and practical usability of the generated reports based on assessment by an experienced pathologist. Furthermore, the quality of the learned representations was evaluated using image-to-text and text-to-image retrieval. For both tasks, we focus on the performance differences for common nevi in comparison to the set of other melanocytic lesion subtypes. (3) We make all code and model parameters publicly available upon acceptance.

\section{Materials and Methods}
\subsection{Dataset}
The experiments in this work were performed using a retrospectively collected dataset from the digital archive of the Department of Pathology at the University Medical Center Utrecht, the Netherlands. The dataset consisted of melanocytic lesion cases accessioned between January 1, 2013, and December 31, 2020. The curation process of the dataset was described in more detail in prior work~\cite{lucassen2024artificial}. Per case, all unique, H\&E-stained WSIs and the corresponding pathology report were included after de-identification. The study was conducted in compliance with the hospital’s research ethics committee guidelines. Cases from patients who opted out of the use of their data for the purpose of research were excluded. 

To prepare the pathology reports for vision-language model development, we developed a custom preprocessing pipeline~\cite{lucassen2024preprocessing}. In brief, the reports were first translated from Dutch to English and afterwards segmented into subsentences based on the information content to enable the selection of specific information. Both tasks were performed using pretrained language models, finetuned for the respective tasks. Part of the findings that are regularly described in pathology reports cannot be derived or are difficult to derive directly from H\&E-stained WSIs, examples of which are results from immunohistochemical stains or molecular tests, potentially causing a model to generate unverifiable statements if trained on this data. To limit this risk, we only included observations related to the cell and tissue appearances written based on the H\&E-stained WSIs and conclusions, such as the diagnosis, from the reports. Although some conclusions were (partially) drawn based on additional diagnostic tests, we consider the conclusion subsentences to be too important to withhold entirely for model development. All cases without any descriptions of cell and tissue patterns based on the H\&E-stained WSIs in the report were excluded from the dataset.

Acquisition of the WSIs was performed using either a ScanScope XT scanner (Aperio, Vista, CA, USA) at 20$\times$ magnification with a resolution of 0.50 \textmu m per pixel (slides scanned before 2016) or a NanoZoomer 2.0-XR scanner (Hamamatsu photonics, Hamamatsu, Shizuoka, Japan) at 40$\times$ magnification with a resolution of 0.23 \textmu m per pixel (slides scanned starting from 2016). Tissue cross-sections and pen markings were segmented in each WSI at 1.25$\times$ magnification using SlideSegmenter~\cite{lucassen2024tissue} to guide the slide tessellation. Non-overlapping tiles of 224$\times$224 pixels were extracted from the WSIs at 20$\times$ magnification. Tiles with identified pen markings or covered by tissue for less than 5\% were excluded.

The dataset comprised of 42,512 H\&E-stained WSIs from 19,645 melanocytic lesions with one report each, acquired from 14,978 patients. Most of these lesions (81.8\%) were benign common nevi. The remaining lesions ranged from benign to intermediate to malignant, including non-common nevi, melanocytomas, and melanomas. In total, the preprocessed reports contained 1,701,127 words across 161,541 sentences. For common nevi, the reports contained an average of 77 words across 7 sentences. In contrast, the reports for all other lesion subtypes contained, on average, 125 words across 12 sentences. The dataset was split on a patient level into sets for training (80\%), validation (10\%), and testing (10\%).

\begin{figure}[t]
    \centering
    \includegraphics[width=0.95\textwidth]{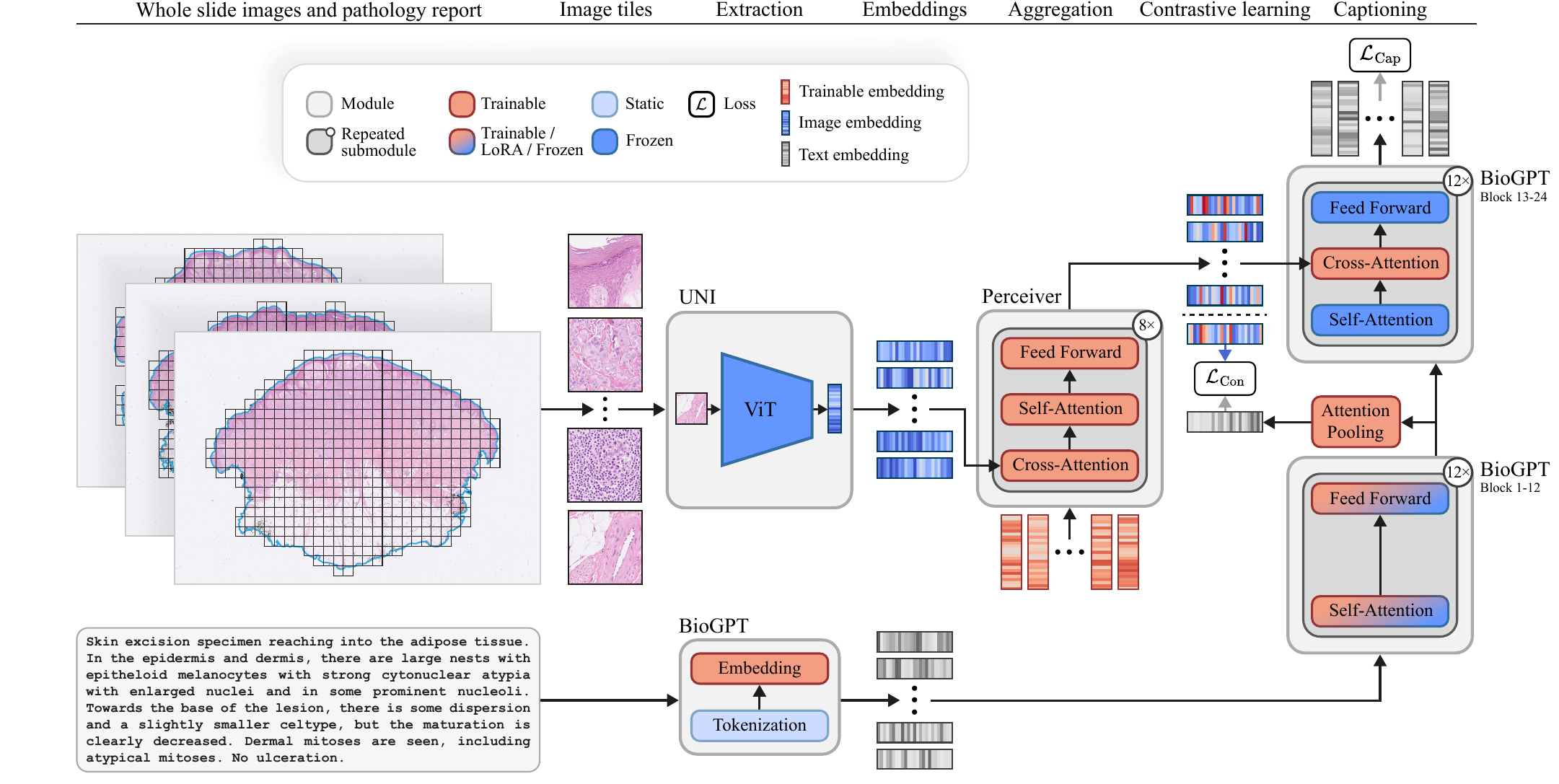}
    \caption{Overview of the vision-language modeling framework. Feature vectors are extracted from all tiles of the tessellated WSIs using UNI~\cite{chen2024towards} and aggregated using the Perceiver~\cite{jaegle2021perceiver}. Preprocessed pathology reports are tokenized and embedded. The vision-language model is trained simultaneously using a contrastive loss and captioning loss.}
    \label{fig:overview}
\end{figure}

\subsection{Vision-Language Model}
The vision-language model we developed follows the design of PRISM~\cite{shaikovski2024prism}, which is an adaptation of the Contrastive Captioner (CoCa) framework~\cite{yu2022coca} that accounts for the large WSIs in computational pathology. An overview of the vision-language model is shown in Fig.~\ref{fig:overview}. 

All extracted WSI tiles for a case were first converted to feature vectors using a frozen, pretrained image encoder. Whereas the Virchow~\cite{vorontsov2024foundation} encoder was used in PRISM, we resorted to the UNI~\cite{chen2024towards} encoder for feature extraction. Aggregation of the 1024-dimensional image feature vectors was performed using the Perceiver~\cite{jaegle2021perceiver}. By leveraging an asymmetric attention mechanism, the Perceiver iteratively distills information from the input feature vectors into a considerably smaller, fixed-length set of 513 trainable embeddings (one of which is used for contrastive training), enabling it to efficiently scale to long input sequences.

The language components of the model were based on BioGPT~\cite{luo2022biogpt}, which has a decoder-only Transformer~\cite{vaswani2017attention} architecture of 24 layers with 347 million parameters and a vocabulary size of 42,384 tokens, pretrained on a biomedical corpus of text. BioGPT was split into a unimodal component (blocks 1-12) and multimodal component (blocks 13-24). An attention-pooling layer was added after the unimodal component for contrastive training. In each layer of the multimodal component, cross-attention layers were inserted to enable interaction between the text embeddings and aggregated image embeddings from the Perceiver.

\subsection{Training}
To limit the required memory during training, only part of the vision-language model components were optimized during training. The Perceiver was trained on the image feature vectors from the frozen UNI tile encoder starting from randomly initialized parameters. The language components were primarily frozen, except for the initial word embedding layer, the attention-pooling layer after the unimodal component, and the cross-attention layers in the multimodal component. Moreover, we experimented with finetuning the unimodal language component both regularly and using Low Rank Adaption (LoRA)~\cite{hu2022lora}.

Following CoCa, the model was trained using a contrastive loss $\mathcal{L}_{\text{Con}}$ and a captioning loss $\mathcal{L}_{\text{Cap}}$. The contrastive loss is optimized by maximizing the similarity between matching pairs of image and text embeddings, while minimizing the similarity between all unmatching pairs of image and text embeddings:
\begin{equation}
    \mathcal{L}_{\text{Con}} = -\frac{1}{N} \left(\sum_{i}^N \log \frac{\exp(x_i^\top y_i / \tau)}{\sum_{j=1}^N \exp(x_i^\top y_j / \tau)} + \sum_{i}^N \log \frac{\exp(y_i^\top x_i / \tau)}{\sum_{j=1}^N \exp(y_i^\top x_j / \tau)} \right)
\end{equation}
where ($x_i$, $y_i$) is the $i$-th matching pair in the batch of $L^2$-normalized image and text embeddings. $N$ is the batch size and $\tau$ is a trainable temperature parameter.

The captioning loss is optimized by minimizing the cross-entropy of the paired text $\mathbf{y}$ using the factorized joint distribution with teacher-forcing~\cite{williams1989learning} to parallelize computation:
\begin{equation}
    \mathcal{L}_{\text{Cap}} = -{\frac{1}{T}\sum_{t=1}^{T}\log P(y_t|\mathbf{y}_{1:t-1}, \mathbf{x})}
\end{equation}
where $y_t$ is the $t$-th token of the report $\mathbf{y}$ with length $T$, $\mathbf{y}_{1:t-1}$ represents all tokens preceding the current token, and $\mathbf{x}$ represents the image embeddings.

The model was trained on the sum of $\mathcal{L}_{\text{Con}}$ and $\mathcal{L}_{\text{Cap}}$ (weighted by a factor of 2) for 30 epochs using bfloat16 precision with a global batch size of 64. Model parameters were updated using a learning rate of 1\,$\cdot$\,10\,\textsuperscript{-4} with a cosine learning rate scheduler and 600 warmup steps. The AdamW~\cite{loshchilov2019decoupled} optimization algorithm ($\beta_1$~=~0.9, $\beta_2$~=~0.999) was used with weight decay equal to 1\,$\cdot$\,10\,\textsuperscript{-6}. Subsets of 100,000 image tiles were randomly sampled during training to reduce the peak in memory usage for the less than 2\% of cases with more tiles available. No tile limit was used during inference. The final model parameters were selected based on the epoch with the lowest validation loss.

\section{Results}
\subsection{Report Generation Performance}
To evaluate the quality of the generated reports, we performed a reader study. A pathologist (G.B.) with experience in dermatopathology was recruited to independently evaluate two reports for a total of 50 cases. These cases were randomly selected from the test set with stratification based on the diagnosis (25 common nevi and 25 lesions of various other subtypes) to cover both common and rare melanocytic lesions. For all selected cases, the generated report and the original report written by a pathologist as part of routine clinical practice were included. The evaluation consisted of counting factual errors, unverifiable statements, omissions of information, and repeated phrases, as well as scoring the report on a scale from 1 to 5. A score of 1 means that the report is mostly inaccurate, offering no useful starting point, whereas a score of 5 indicates that the report is highly accurate with minimal to no adjustments required for practical use. The pathologist only had access to the WSIs during the reader study. To prevent bias in the evaluation, reports were randomly ordered per case and the pathologist was blinded from the origin of the report.

\begin{table}[t]
\centering
\caption{Results of the reader study with blinded evaluation by a pathologist are presented as the mean and standard deviation. The score reflects the overall accuracy and usability of the reports on a 1-5 scale. The count of unverifiable statements is not applicable to pathologist-written reports.}
\begin{adjustbox}{width=\textwidth}
\begin{tabular}{@{}cclcccccccc@{}}
\toprule \toprule
 & Data            & Written by    & ~~ & \multicolumn{4}{c}{Error count per report}  & ~~ & Score         & \\
 &                 &               &    & ~Factual~                    & Unverifiable~                 & Omission~                    & Repetition                  & ~~ &                            & \\ \midrule
 & All             & Model         &    & 1.4{\scriptsize~$\pm$~2.1}   & 0.0{\scriptsize~$\pm$~0.0}    & 0.7{\scriptsize~$\pm$~1.0}   & 1.6{\scriptsize~$\pm$~2.9}  & ~~ & 3.7{\scriptsize~$\pm$~1.2} & ~ \\
 & ($N$\,=\,50)    & Pathologist   &    & 0.5{\scriptsize~$\pm$~0.8}   & --                            & 0.2{\scriptsize~$\pm$~0.5}   & 0.3{\scriptsize~$\pm$~1.1}  & ~~ & 4.6{\scriptsize~$\pm$~0.6} & ~ \\ \rule{0pt}{3ex}    
 & Common nevi~    & Model         &    & 0.4{\scriptsize~$\pm$~0.6}   & 0.0{\scriptsize~$\pm$~0.0}    & 0.3{\scriptsize~$\pm$~0.5}   & 0.2{\scriptsize~$\pm$~0.7}  & ~~ & 4.5{\scriptsize~$\pm$~0.8} & ~ \\
 & ($N$\,=\,25)    & Pathologist   &    & 0.4{\scriptsize~$\pm$~0.7}   & --                            & 0.1{\scriptsize~$\pm$~0.3}   & 0.0{\scriptsize~$\pm$~0.0}  & ~~ & 4.6{\scriptsize~$\pm$~0.6} & ~ \\ \rule{0pt}{3ex}  
 & Other lesions   & Model         &    & 2.5{\scriptsize~$\pm$~2.5}   & 0.0{\scriptsize~$\pm$~0.0}    & 1.1{\scriptsize~$\pm$~1.1}   & 2.9{\scriptsize~$\pm$~3.6}  & ~~ & 3.0{\scriptsize~$\pm$~1.1} & ~ \\
 & ($N$\,=\,25)    & Pathologist   &    & 0.5{\scriptsize~$\pm$~0.9}   & --                            & 0.3{\scriptsize~$\pm$~0.6}   & 0.6{\scriptsize~$\pm$~1.6}  & ~~ & 4.5{\scriptsize~$\pm$~0.7} & ~ \\
\bottomrule \bottomrule
\end{tabular}
\end{adjustbox}
\label{tab:reader_study}
\end{table}

\begin{figure}[t]
    \centering
    \includegraphics[width=\textwidth]{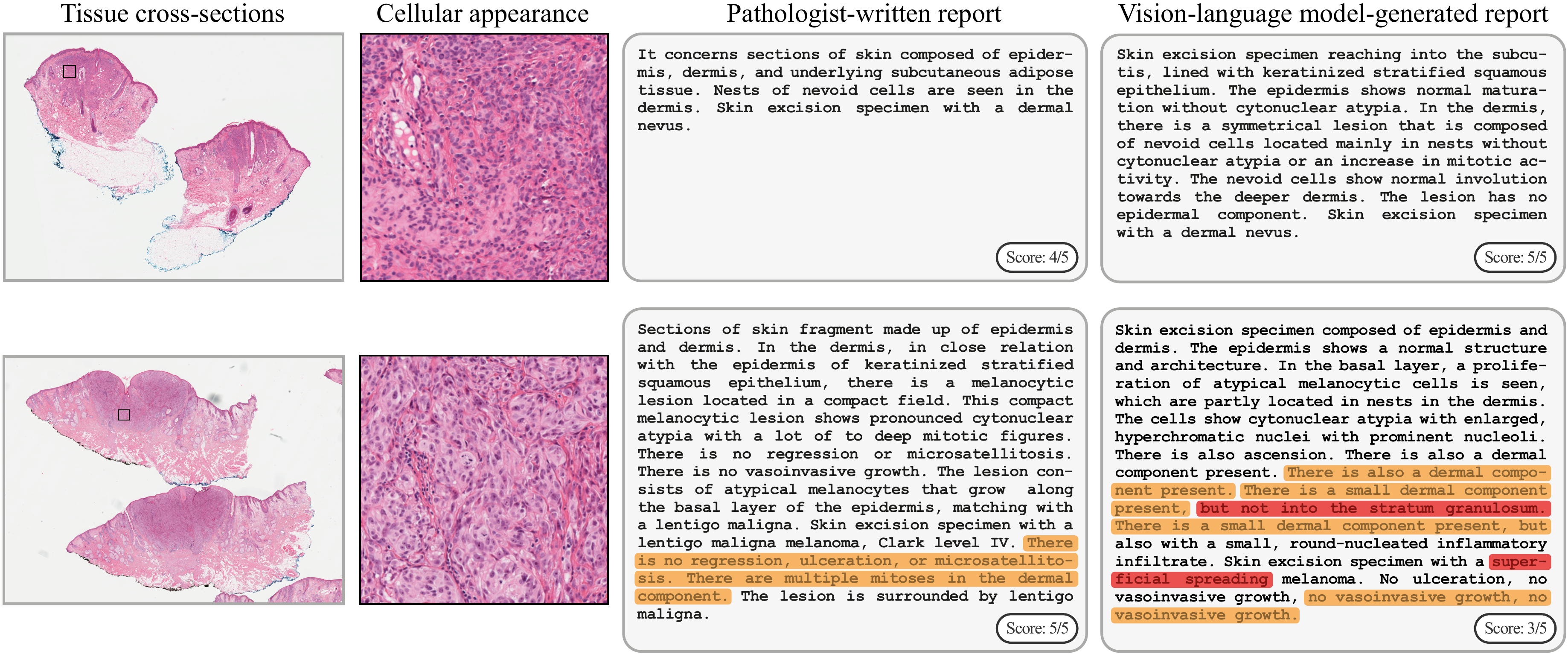}
    \caption{Two example cases from the reader study. Factual errors are highlighted in red and repeated phrases in orange.}
    \label{fig:result}
\end{figure}

The mean and standard deviation of the error counts and quality scores for the reader study are shown in Table~\ref{tab:reader_study}. Two example cases can be seen in Fig.~\ref{fig:result}. On the complete subset, the reports generated by the vision-language model scored an average of 3.7\,($\pm$\,1.2) out of 5, whereas the reports written by pathologists scored a 4.6\,($\pm$\,0.6) out of 5. In terms of performance, there are clear differences between cases with common nevi and other melanocytic lesions. For the common nevi, the model-generated reports scored comparably to the pathologist-written reports, although slightly more factual errors, repeated phrases, and omissions of important information were recorded. Larger differences were seen for the other melanocytic lesions. The generated reports scored a 3.0\,($\pm$\,1.1) out of 5, on average, which is substantially below the 4.5\,($\pm$\,0.7) out of 5 for the original reports. This is also reflected by the larger number of factual errors, omissions, and repetitions. No unverifiable sentences were identified in the generated reports.

\begin{table}[t]
\centering
\caption{Results for image-to-text matching based on the cases in the independent test set. Note that lower scores represent better performance for the rank.}
\setlength{\tabcolsep}{3pt}
\begin{adjustbox}{width=\textwidth,center}
\begin{tabular}{@{}cccccccccc@{}}
\toprule \toprule
 & Data                                    &  & \multicolumn{3}{c}{Recall at $k$}                     &  & \multicolumn{2}{c}{Rank} &  \\
 &                                         &  & $k$\,=\,1   & $k$\,=\,5   & $k$\,=\,10  &  & Mean       & Median      &  \\ \midrule
 & All                    &  & 0.094        & 0.268        & 0.375        &  & 66.7        & 21           &  \\
 & ($N$\,=\,1,969)        &  & {\scriptsize(0.082-0.107)} & {\scriptsize(0.250-0.288)} & {\scriptsize(0.355-0.396)} &  & {\scriptsize(62.0-72.2)}  &  {\scriptsize(19-23)}        &  \\ \rule{0pt}{3ex}
 & Common nevi            &  & 0.072        & 0.210        & 0.310        &  & 78.1        & 28            &  \\
 &($N$\,=\,1,592)         &  & {\scriptsize(0.049-0.073)} & {\scriptsize(0.181-0.222)} & {\scriptsize(0.281-0.326)} &  & {\scriptsize(72.6-84.9)} & {\scriptsize(26-33)}          &  \\ \rule{0pt}{3ex}
 & Other lesions          &  & 0.193        & 0.509        & 0.646        &  & 18.8           & 5            &  \\
 & ($N$\,=\,377)          &  & {\scriptsize(0.126-0.198)} & {\scriptsize(0.429-0.534)} & {\scriptsize(0.582-0.678)} &  & {\scriptsize(15.0-24.7)} & {\scriptsize(5-7)}            &  \\
 \bottomrule \bottomrule
\end{tabular}
\end{adjustbox}
\label{Tab:I2T}
\end{table}

\begin{table}[t]
\centering
\caption{Results for text-to-image matching based on the cases in the independent test set. Note that lower scores represent better performance for the rank.}
\setlength{\tabcolsep}{3pt}
\begin{adjustbox}{width=\textwidth,center}
\begin{tabular}{@{}cccccccccc@{}}
\toprule \toprule
 & Data                                    &  & \multicolumn{3}{c}{Recall at $k$}                     &  & \multicolumn{2}{c}{Rank} &  \\
 &                                         &  & $k$\,=\,1   & $k$\,=\,5   & $k$\,=\,10  &  & Mean       & Median      &  \\ \midrule
 & All                    &  & 0.090        & 0.247        & 0.373        &  & 67.4           & 20            &  \\
 & ($N$\,=\,1,969)         &  & {\scriptsize(0.079-0.103)} & {\scriptsize(0.229-0.266)} & {\scriptsize(0.352-0.396)} &  & {\scriptsize(61.9-72.6)} & {\scriptsize(18-22)} &  \\ \rule{0pt}{3ex}
 & Common nevi            &  & 0.069        & 0.198        & 0.311        &  & 78.6           & 29            &  \\
 &($N$\,=\,1,592)         &  & {\scriptsize(0.045-0.069)} & {\scriptsize(0.175-0.212)} & {\scriptsize(0.281-0.329)} &  & {\scriptsize(73.1-85.5)} & {\scriptsize(26-33)} &  \\ \rule{0pt}{3ex}
 & Other lesions          &  & 0.185        & 0.456        & 0.635        &  & 20.0           & 7            &  \\
 & ($N$\,=\,377)           &  & {\scriptsize(0.121-0.193)} & {\scriptsize(0.397-0.493)} & {\scriptsize(0.568-0.670)} &  &  {\scriptsize(15.9-26.3)} & {\scriptsize(6-8)} &  \\
 \bottomrule \bottomrule
\end{tabular}
\end{adjustbox}
\label{Tab:T2I}
\end{table}

\subsection{Retrieval Performance}
The quality of the learned representations was assessed based on the retrieval performance across modalities using the cases from the independent test set ($N$\,=\,1,969). This evaluation measures the extent to which pathology reports can be matched to their corresponding WSIs (and vice versa) based on the similarity of image and text representations. Performance on the retrieval tasks was expressed in terms of the recall at $k$ (i.e., the proportion of cases for which the matching item is in the top $k$ retrieved items), as well as the mean and median rank. To estimate 95\% confidence intervals (CIs), bootstrapping ($R$~=~1,000 samples) was performed using the percentile method. The set of items to be retrieved was not sampled during the bootstrapping procedure to prevent matching conflicts for duplicates. Results for the lesion subsets represent the average of the global retrieval performance with all cases available.

The best mean rank of retrieved images matched to texts was achieved using frozen BioGPT layers and equal to 66.7 (95\% CI, 62.0-72.2), closely followed by LoRA finetuning with 68.4 (95\% CI, 63.6-73.6), both outperforming regular finetuning with 82.8 (95\% CI, 77.2-89.1). Similar trends were seen across the median rank and recall scores, as well as for text-to-image matching. More detailed results for the model with frozen BioGPT layers are shown in Table~\ref{Tab:I2T} for image-to-text matching and in Table~\ref{Tab:T2I} for text-to-image matching. In approximately 37\% of the test set cases, the matching image or report was retrieved as part of the 10 best matching counterparts. The retrieval performance was substantially worse for common nevi than for the other melanocytic lesion subtypes.

\section{Discussion and Conclusion}
In this work, we developed a vision-language model specifically for the pathology domain of cutaneous melanocytic lesions. The model performance was assessed based on two potential use cases for a digital pathology department: report generation and content-based case retrieval.

Reports generated by the model for common nevi were scored comparably to pathologist-written reports, whereas the generated reports for other melanocytic lesion subtypes received a lower average score. We expect that this is because reports for common nevi tend to be somewhat shorter, more similar, and well-represented in the dataset. The fact that common nevi form the majority of all melanocytic lesion cases, in combination with the strong model performance, makes this a promising direction to reduce pathologists' workload in clinical practice. The presence of repeated phrases, which was especially observed for the more uncommon melanocytic lesion subtypes, has not been described in related research on pathology-specific vision-language modeling. This might be due to the more challenging task of writing coherent reports that are longer than one or two sentences, as well as the limited representation of rare subtypes. While the reader study showed clear patterns, recruiting multiple readers and selecting a larger set of cases would be important for a more comprehensive analysis.

In contrast to the report generation, the cross-modal retrieval evaluation showed substantially worse results for the common nevi in comparison to the other melanocytic lesions. This can likely be attributed to the same factors, as it is more difficult to retrieve the original counterpart based on shorter reports from a larger set of similar cases, of which many cases may fit the description or appearance reasonably if not exactly. While the study by Ding~\textit{et al.}~\cite{ding2024multimodal} reported much higher recall values using a pan-cancer dataset, their retrieval experiment considered all cases with the same diagnosis a correct match. A system that combines uni- or cross-modal similarity search with diagnostic codes and clinical metadata might improve retrieval performance and could facilitate more diverse queries.

In conclusion, this study demonstrates that vision-language modeling has potential for automated pathology report generation, particularly for common nevi. While generating reports for more uncommon melanocytic lesion subtypes revealed to be more challenging, the cross-modal retrieval performance was substantially better for these cases.

\begin{credits}
\subsubsection{\ackname} This research was financially supported by the Hanarth Foundation.
\subsubsection{\discintname} The authors declare no competing interests.
\end{credits}

\bibliographystyle{splncs04}
\bibliography{mybibliography}

\begin{thebibliography}{10}
\providecommand{\url}[1]{\texttt{#1}}
\providecommand{\urlprefix}{URL }
\providecommand{\doi}[1]{https://doi.org/#1}

\bibitem{ahmed2024pathalign}
Ahmed, F., Sellergen, A., Yang, L., Xu, S., Babenko, B., Ward, A., Olson, N., Mohtashamian, A., Matias, Y., Corrado, G.S., Duong, Q., Webster, D.R., Shetty, S., Golden, D., Liu, Y., Steiner, D.F., Wulczyn, E.: Path{A}lign: {A} vision–language model for whole slide images in histopathology. In: Proceedings of the MICCAI Workshop on Computational Pathology. Proceedings of Machine Learning Research, vol.~254, pp. 72--108 (2024)

\bibitem{berbis2023computational}
Berb{\'\i}s, M.A., McClintock, D.S., Bychkov, A., Van~der Laak, J., Pantanowitz, L., Lennerz, J.K., Cheng, J.Y., Delahunt, B., Egevad, L., Eloy, C., et~al.: Computational pathology in 2030: {A} {D}elphi study forecasting the role of {AI} in pathology within the next decade. EBioMedicine  \textbf{88} (2023)

\bibitem{bray2024global}
Bray, F., Laversanne, M., Sung, H., Ferlay, J., Siegel, R.L., Soerjomataram, I., Jemal, A.: Global cancer statistics 2022: {GLOBOCAN} estimates of incidence and mortality worldwide for 36 cancers in 185 countries. CA: a cancer journal for clinicians  \textbf{74}(3),  229--263 (2024)

\bibitem{chen2024towards}
Chen, R.J., Ding, T., Lu, M.Y., Williamson, D.F.K., Jaume, G., Song, A.H., Chen, B., Zhang, A., Shao, D., Shaban, M., Williams, M., Oldenburg, L., Weishaupt, L.L., Wang, J.J., Vaidya, A., Le, L.P., Gerber, G., Sahai, S., Williams, W., Mahmood, F.: Towards a general-purpose foundation model for computational pathology. Nature Medicine  \textbf{30}(3),  850--862 (2024)

\bibitem{ding2024multimodal}
Ding, T., Wagner, S.J., Song, A.H., Chen, R.J., Lu, M.Y., Zhang, A., Vaidya, A.J., Jaume, G., Shaban, M., Kim, A., Williamson, D.F.K., Chen, B., Almagro-Perez, C., Doucet, P., Sahai, S., Chen, C., Koruma, D., Kawabe, S., Gerber, G., Peng, T., Le, L.P., Mahmood, F.: Multimodal whole slide foundation model for pathology. arXiv preprint arXiv:2411.19666  (2024)

\bibitem{hu2022lora}
Hu, E.J., Shen, Y., Wallis, P., Allen-Zhu, Z., Li, Y., Wang, S., Wang, L., Chen, W.: Lo{RA}: {L}ow-rank adaptation of large language models. In: International Conference on Learning Representations (2022)

\bibitem{huang2023visual}
Huang, Z., Bianchi, F., Yuksekgonul, M., Montine, T.J., Zou, J.: A visual--language foundation model for pathology image analysis using medical {T}witter. Nature medicine  \textbf{29}(9),  2307--2316 (2023)

\bibitem{jaegle2021perceiver}
Jaegle, A., Gimeno, F., Brock, A., Vinyals, O., Zisserman, A., Carreira, J.: Perceiver: {G}eneral perception with iterative attention. In: International conference on machine learning. pp. 4651--4664 (2021)

\bibitem{ji2023survey}
Ji, Z., Lee, N., Frieske, R., Yu, T., Su, D., Xu, Y., Ishii, E., Bang, Y.J., Madotto, A., Fung, P.: Survey of hallucination in natural language generation. ACM Computing Surveys  \textbf{55}(12),  1--38 (2023)

\bibitem{vanderLaak2021}
Van~der Laak, J., Litjens, G., Ciompi, F.: Deep learning in histopathology: {T}he path to the clinic. Nature Medicine  \textbf{27},  775--784 (2021)

\bibitem{li2023blip}
Li, J., Li, D., Savarese, S., Hoi, S.: {BLIP}-2: {B}ootstrapping language-image pre-training with frozen image encoders and large language models. In: International conference on machine learning. pp. 19730--19742 (2023)

\bibitem{liu2024visual}
Liu, H., Li, C., Wu, Q., Lee, Y.J.: Visual instruction tuning. In: Advances in neural information processing systems. vol.~36 (2024)

\bibitem{loshchilov2019decoupled}
Loshchilov, I., Hutter, F.: Decoupled weight decay regularization. In: Proceedings of the International Conference on Learning Representations (2019)

\bibitem{lott2018population}
Lott, J.P., Boudreau, D.M., Barnhill, R.L., Weinstock, M.A., Knopp, E., Piepkorn, M.W., Elder, D.E., Knezevich, S.R., Baer, A., Tosteson, A.N., et~al.: Population-based analysis of histologically confirmed melanocytic proliferations using natural language processing. JAMA dermatology  \textbf{154}(1),  24--29 (2018)

\bibitem{lu2024visual}
Lu, M.Y., Chen, B., Williamson, D.F.K., Chen, R.J., Liang, I., Ding, T., Jaume, G., Odintsov, I., Le, L.P., Gerber, G., Parwani, A.V., Zhang, A., Mahmood, F.: A visual-language foundation model for computational pathology. Nature Medicine  \textbf{30}(3),  863--874 (2024)

\bibitem{lucassen2024tissue}
Lucassen, R.T., Blokx, W.A.M., Veta, M.: Tissue cross-section and pen marking segmentation in whole slide images. In: Proceedings of SPIE 12933, Medical Imaging 2024: Digital and Computational Pathology. vol. 12933 (2024)

\bibitem{lucassen2024preprocessing}
Lucassen, R.T., Van~de Luijtgaarden, T., Moonemans, S.P.J., Blokx, W.A.M., Veta, M.: Preprocessing pathology reports for vision-language model development. In: Proceedings of the MICCAI Workshop on Computational Pathology. Proceedings of Machine Learning Research, vol.~254, pp. 61--71 (2024)

\bibitem{lucassen2024artificial}
Lucassen, R.T., Stathonikos, N., Breimer, G.E., Veta, M., Blokx, W.A.: Artificial intelligence-based triaging of cutaneous melanocytic lesions. arXiv preprint arXiv:2410.10509  (2024)

\bibitem{luo2022biogpt}
Luo, R., Sun, L., Xia, Y., Qin, T., Zhang, S., Poon, H., Liu, T.Y.: Bio{GPT}: {G}enerative pre-trained transformer for biomedical text generation and mining. Briefings in bioinformatics  \textbf{23}(6),  1--11 (2022)

\bibitem{metter2019trends}
Metter, D.M., Colgan, T.J., Leung, S.T., Timmons, C.F., Park, J.Y.: Trends in the {US} and {C}anadian pathologist workforces from 2007 to 2017. JAMA network open  \textbf{2}(5) (2019)

\bibitem{radford2021learning}
Radford, A., Kim, J.W., Hallacy, C., Ramesh, A., Goh, G., Agarwal, S., Sastry, G., Askell, A., Mishkin, P., Clark, J., Krueger, G., Sutskever, I.: Learning transferable visual models from natural language supervision. In: International conference on machine learning. pp. 8748--8763 (2021)

\bibitem{shaikovski2024prism}
Shaikovski, G., Casson, A., Severson, K., Zimmermann, E., Wang, Y.K., Kunz, J.D., Retamero, J.A., Oakley, G., Klimstra, D., Kanan, C., Hanna, M., Zelechowski, M., Viret, J., Tenenholtz, N., Hall, J., Fusi, N., Yousfi, R., Hamilton, P., Moye, W.A., Vorontsov, E., Liu, S., Fuchs, T.J.: {PRISM}: {A} multi-modal generative foundation model for slide-level histopathology. arXiv preprint arXiv:2405.10254  (2024)

\bibitem{vaswani2017attention}
Vaswani, A., Shazeer, N., Parmar, N., Uszkoreit, J., Jones, L., Gomez, A.N., Kaiser, {\L}., Polosukhin, I.: Attention is all you need. In: Advances in Neural Information Processing Systems. vol.~30 (2017)

\bibitem{vorontsov2024foundation}
Vorontsov, E., Bozkurt, A., Casson, A., Shaikovski, G., Zelechowski, M., Severson, K., Zimmermann, E., Hall, J., Tenenholtz, N., Fusi, N., Yang, E., Mathieu, P., van Eck, A., Lee, D., Viret, J., Robert, E., Wang, Y.K., Kunz, J.D., Lee, M.C.H., Bernhard, J.H., Godrich, R.A., Oakley, G., Millar, E., Hanna, M., Wen, H., Retamero, J.A., Moye, W.A., Yousfi, R., Kanan, C., Klimstra, D.S., Rothrock, B., Liu, S., Fuchs, T.J.: A foundation model for clinical-grade computational pathology and rare cancers detection. Nature medicine pp. 1--12 (2024)

\bibitem{williams1989learning}
Williams, R.J., Zipser, D.: A learning algorithm for continually running fully recurrent neural networks. Neural computation  \textbf{1}(2),  270--280 (1989)

\bibitem{yu2022coca}
Yu, J., Wang, Z., Vasudevan, V., Yeung, L., Seyedhosseini, M., Wu, Y.: Co{C}a: {C}ontrastive captioners are image-text foundation models. Transactions on Machine Learning Research  (2022)

\end{thebibliography}

\end{document}